# Dipole and Quadrupole Moments in Image Processing

A. Sparavigna
Dipartimento di Fisica, Politecnico di Torino, Torino, Italy
amelia.sparavigna@polito.it

**Abstract**: This paper proposes an algorithm for image processing, obtained by adapting to image maps the definitions of two well-known physical quantities. These quantities are the dipole and quadrupole moments of a charge distribution. We will see how it is possible to define dipole and quadrupole moments for the gray-tone maps and apply them in the development of algorithms for edge detection.

**Keywords:** Image analysis, edge localization.

## 1. Introduction
A well-known concept used in condensed matter physics and optics, that of coherence length, has been recently adapted for texture analysis of images. According to this physical concept, we developed an algorithm able to compare local and global behaviors of image maps [1-4]. By means of this algorithm we characterized irregular textures, where Fourier or wavelets approaches are less useful and statistical evaluations are preferred [5,6].
In this paper again, we propose a new approach for image processing, developed by adaptation of two other physical quantities[1]. We will show how the dipole and quadrupole moments of charge distribution turn out to be very interesting quantities for image pattern recognition. After defining the dipole and quadrupole moments for image maps, we will use them in the edge detection.

## 2. Dipoles and quadrupole moments.
A general distribution of electric charges can be described by its net charge, by its dipole and quadrupole moments, and higher order moments [7]. Usually, for a good description of charge distribution, it is enough the development till the quadrupole moment. The dipole moment for a pair of opposite charges of magnitude $q$ is defined as the magnitude of the charge times the distance between them. The dipole moment is a vector, the direction of which is toward the positive charge.
Dipoles and quadrupoles are used for charge distribution in nuclei, atoms and molecules, where the effects of charge separation are measurable but distances between charges are too small to be measured. Dipoles are fundamental for studying dielectrics, liquids and liquid crystals; sometimes quadrupoles are useful in liquid crystals. The quadrupole moment is usually represented by a traceless tensor.

---

[1] Let us remember that a physical quantity is a property that can be quantified. That is, it is possible to give a quantitative attribute, which exists in a range of magnitudes, and can therefore be measured.

Consider a collection of $N$ particles with charges $q_n$ and position vectors $\vec{r}_n$. The physical quantity defining the dipole vector is given by:

$$\vec{p} = \sum_{n=1}^{N} q_n \vec{r}_n \tag{1}$$

For a distribution of charges in a plane $(x, y)$:

$$p_x = \sum_{n=1}^{N} q_n x_n \quad ; \quad p_y = \sum_{n=1}^{N} q_n y_n \tag{2}$$

where $x_n, y_n$ are the Cartesian components of the position vector of each charge. We are assuming a distribution of charges in $(x, y)$-plane, because we want to apply these moments to a bidimensional image. In fact, the proposed algorithm can be easily extended to the study of three-dimensional objects.

The traceless tensor describing the quadrupole moment of a bidimensional discrete distribution of charges (or masses, for example) is defined as:

$$Q_{xx} = \sum_{n=1}^{N} q_n \left( 2 x_n x_n - r_n^2 \right)$$

$$Q_{yy} = \sum_{n=1}^{N} q_n \left( 2 y_n y_n - r_n^2 \right) \tag{3}$$

$$Q_{xy} = \sum_{n=1}^{N} q_n \left( 2 x_n y_n \right)$$

where $r_n^2 = x_n^2 + y_n^2$.

**3. Dipole and quadrupoles in images.**

In the case of images, we have a quantity which can play the role of a charge distribution and this quantity is the image bitmap $b$. For simplicity let us assume a gray-scale color coding. The bitmap representation of an image consists of a function which yields the brightness of each point within a specific width and height range:

$$b: D \to B \qquad (4)$$

with $D = I_h \times I_w$, where $I_h = \{1,2,\ldots,h\} \subset \mathbb{N}$, $I_w = \{1,2,\ldots,w\} \subset \mathbb{N}$ and $B = \{0,1,\ldots,255\} \subset \mathbb{N}$. Of course we can evaluate a dipole and quadruple moments for the overall image in the following way:

$$p_x = \frac{1}{hw} \sum_{i=1}^{h} \sum_{j=1}^{w} b(i,j) x_{ij} \; ; \; p_y = \frac{1}{hw} \sum_{i=1}^{h} \sum_{j=1}^{w} b(i,j) y_{ij} \qquad (5)$$

where $x_{ij}, y_{ij}$ are the coordinates of pixel at position $(i,j)$ in the image frame. Explicitly, these coordinates can be simply identified with pixel positions: $x_{ij} = i, y_{ij} = j$. A traceless quadrupole moment tensor can be defined as:

$$Q_{xx} = \frac{1}{hw} \sum_{i=1}^{h} \sum_{j=1}^{w} b(i,j)\left(2 x_{ij} x_{ij} - r_{ij}^2\right)$$

$$Q_{yy} = \frac{1}{hw} \sum_{i=1}^{h} \sum_{j=1}^{w} b(i,j)\left(2 y_{ij} y_{ij} - r_{ij}^2\right) \qquad (6)$$

$$Q_{xx} = \frac{1}{hw} \sum_{i=1}^{h} \sum_{j=1}^{w} b(i,j)\left(2 x_{ij} y_{ij}\right)$$

where $r_{ij}^2 = x_{ij}^2 + y_{ij}^2$. If we consider all the image frame, we have the distribution of gray tones lumped in two parameters, but this is not interesting for processing the image under consideration.

Instead to consider the whole image, let us evaluate dipole and quadrupole moments on a neighborhood of each pixel. As usual, the pixel position in the image map is given by indices $(i,j)$. The neighborhood consists of all pixels with indices contained in the two following intervals $I_i = [i - \Delta i, i + \Delta i]$ and $I_j = [j - \Delta j, j + \Delta j]$. The local average brightness is defined as:

$$M(i,j) = \frac{1}{4 \Delta i \, \Delta j} \sum_{k \in I_i} \sum_{l \in I_j} b(k,l) \qquad (7)$$

Now, a pixel in this local neighborhood can have a "charge", that can be positive or negative, if we define the "charge" as $q(i,j) = b(i,j) - M(i,j)$.
Local dipoles and quadrupoles are then given by:

$$p_x(i,j) = \frac{1}{4\Delta i \, \Delta j} \sum_{k \in I_i} \sum_{l \in I_j} q(k,l) x_{kl}$$

$$p_y(i,j) = \frac{1}{4\Delta i \, \Delta j} \sum_{k \in I_i} \sum_{l \in I_j} q(k,l) y_{kl} \qquad (8)$$

and

$$Q_{xx}(i,j) = \frac{1}{4\Delta i \, \Delta j} \sum_{k \in I_i} \sum_{l \in I_j} q(k,l)\left(2 x_{kl} x_{kl} - r_{kl}^2\right)$$

$$Q_{yy}(i,j) = \frac{1}{4\Delta i \, \Delta j} \sum_{k \in I_i} \sum_{l \in I_j} q(k,l)\left(2 y_{kl} y_{kl} - r_{kl}^2\right) \qquad (9)$$

$$Q_{xy}(i,j) = \frac{1}{4\Delta i \, \Delta j} \sum_{k \in I_i} \sum_{l \in I_j} q(k,l)\left(x_{kl} y_{kl}\right)$$

We can prepare two image maps representing the dipole distribution and the quadrupole distribution respectively. Of course, these quantities are vectors and tensors, and a complete description is far from being simple. If we consider just the magnitude of dipole vector and the absolute value of quadrupole determinant, we have two scalar quantities which can be easily associated with two gray tone maps. These two scalars are:

$$P(i,j) = \left(p_x^2(i,j) + p_y^2(i,j)\right)^{1/2}$$

$$Q(i,j) = \left| Q_{xx}(i,j) Q_{yy}(i,j) - Q_{xy}^2(i,j) \right| \qquad (10)$$

Let us evaluate the two maximum values of $P(i,j)$ and $Q(i,j)$ in the total image frame and call them $P_{Max}, Q_{Max}$. To represent with an image map the distribution of dipoles or quadrupoles, we associate with each pixel a gray tone as follows:

$$b_P(i,j) = 255 \left(\frac{P(i,j)}{P_{Max}}\right)^{\alpha} \qquad (11)$$

$$b_Q(i,j) = 255 \left(\frac{Q(i,j)}{Q_{Max}}\right)^{\beta} \qquad (12)$$

where exponents $\alpha, \beta$ are properly adjusted to enhance the visibility of respective maps. We will see in the next section maps with $\alpha = 1/2, \beta = 1/4$.

**4. An edge detector.**
Image dipole and quadrupole moments describe the distribution of tones, as the dipoles and quadrupoles describe the distribution of charges. We told, for what concerns charges, that dipole and quadrupole moments can give huge effects also in the case of small distances among charges. It is then natural to test our image dipole and quadrupole moments in image edge detection, where we have to detect abrupt changes of brightness in small regions of the image frame.

A sharp brightness change corresponds to changes in properties of physical objects. Discontinuities in image brightness are likely to correspond to discontinuities in depth or in surface orientation and changes in material properties. Moreover they can correspond to variations in scene illumination. Applying an edge detector to an image map gives a set of curves describing the boundaries of objects. But these boundaries can be used to represent the real objects. For this reason, the edge detection is very important to reduce the amount of data, while preserving the relevant structural properties of an image.

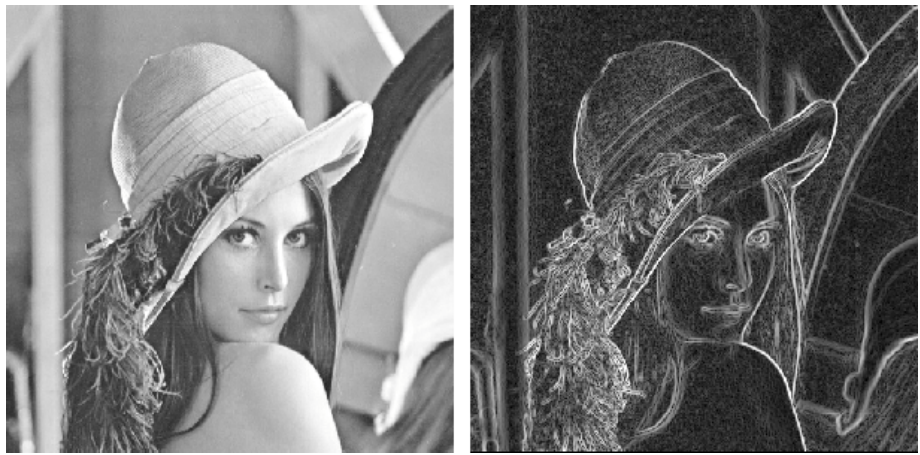

Fig.1 The original image on the left. On the right, the map obtained evaluating the magnitude of local dipole moments, according to Eqs.10 and 11. Note that this map is giving the image edges.

Let us apply the dipole approach to a test image. The dipole moment is evaluated on the smallest possible neighborhood, that is the neighborhood with $2 \times 2$ pixels. Fig.1 shows the original image and the map representing the magnitude of local dipole moments. This map clearly shows the image edges and it is similar to the map which can be obtained using a software such as GIMP.

In the case we apply the quadrupole approach, we obtain the map shown in Fig.2. As in the case of the dipole map, the used neighborhood is that with $2\times 2$ pixels. Fig.2 gives the map representing the absolute value of determinant, map which is obtained by means of Eqs.10 and 12. This map again is clearly showing the edges in the image.

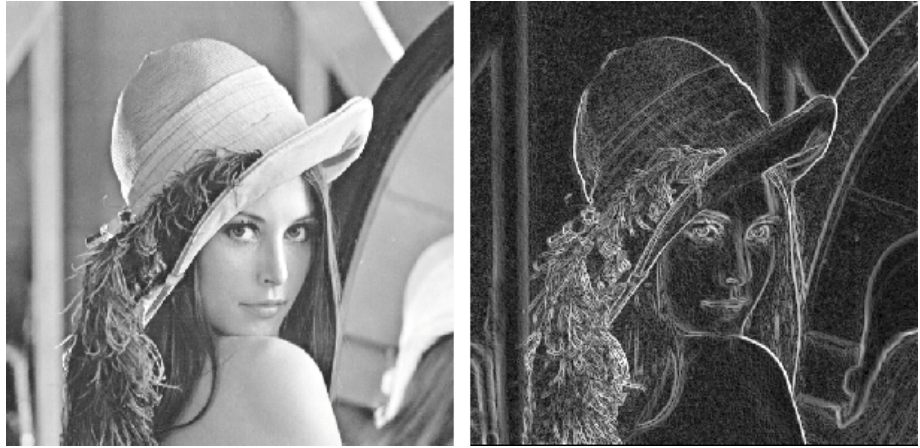

Fig.2 The original image map on the left. On the right, the map obtained evaluating the determinant of local quadrupole moments, according to Eqs.10 and 12. Note that the quadrupole distribution is giving again the edges.

To illustrate the algorithm with another example, let us apply dipole and quadrupole moments to an image of Brodatz album [8]. Results are shown in Figs.3 and 4.

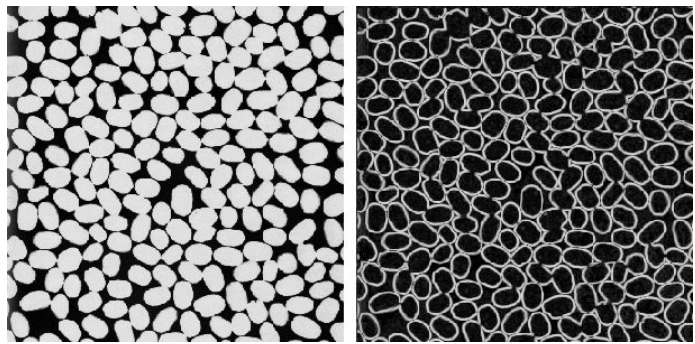

Fig.3 The original image map on the left from the Brodatz album. On the right, the map obtained evaluating the dipole moments, according to Eqs. 10 and 11.

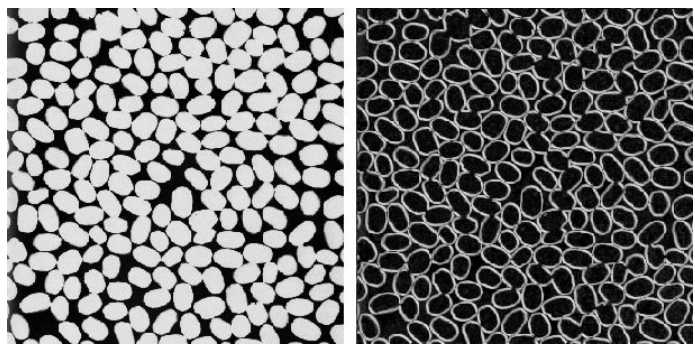

Fig.4 The original image map on the left. On the right, the map obtained evaluating the quadrupole moments, according to Eqs.10 and 12.

**5. Conclusions**
The paper describes an algorithm based on image dipole and quadrupole moments. These moments are defined as in physics are defined dipole and quadrupole moments of a charge distribution. With this algorithm, evaluating the moments on small neighborhoods of each pixel, it is possible to detect the edges in the image frame.